\documentclass[conference]{IEEEtran}
\IEEEoverridecommandlockouts
\usepackage{graphicx}
\usepackage{cite}
\usepackage{picinpar}
\usepackage{amsmath}
\usepackage{url}
\usepackage{flushend}
\usepackage[latin1]{inputenc}
\usepackage{colortbl}
\usepackage{soul}
\usepackage{multirow}
\usepackage{pifont}
\usepackage{color}
\usepackage{alltt}
\usepackage[hidelinks]{hyperref}
\usepackage{enumerate}
\usepackage{siunitx}
\usepackage{epstopdf}
\usepackage{pbox}
\usepackage{amsfonts}
\usepackage{bm}
\usepackage{empheq}
\usepackage[caption=false, font=footnotesize]{subfig}

\title{\LARGE\bf
Vision-Based Dexterous Motion Planning by Dynamic Movement Primitives with Human Hand Demonstration
}
\author{Nuo~Chen,~\IEEEmembership{Student~Member,~IEEE},
        and Ya-Jun~Pan,~\IEEEmembership{Senior~Member,~IEEE}
\thanks{This work was supported by the Natural Sciences and Engineering Research Council (NSERC) and the Government of Nova Scotia, Canada.}
\thanks{Nuo Chen and Ya-Jun Pan are with the Advanced Control and Mechatronics Lab, Department of Mechanical Engineering, Dalhousie University, Halifax, Canada, B3H 4R2. (e-mails: {\tt\small nuo.chen@dal.ca}, {\tt\small yajun.pan@dal.ca}).}
}
\begin{document}
\maketitle
\thispagestyle{empty}
\pagestyle{empty}
\begin{abstract}
This paper proposes a vision-based framework for a 7-degree-of-freedom robotic manipulator, with the primary objective of facilitating its capacity to acquire information from human hand demonstrations for the execution of dexterous pick-and-place tasks. Most existing works only focus on the position demonstration without considering the orientations. In this paper, by employing a single depth camera, MediaPipe is applied to generate the three-dimensional coordinates of a human hand, thereby comprehensively recording the hand's motion, encompassing the trajectory of the wrist, orientation of the hand, and the grasp motion. A mean filter is applied during data pre-processing to smooth the raw data. The demonstration is designed to pick up an object at a specific angle, navigate around obstacles in its path and subsequently, deposit it within a sloped container. The robotic system demonstrates its learning capabilities, facilitated by the implementation of Dynamic Movement Primitives, enabling the assimilation of user actions into its trajectories with different start and end points. Experimental studies are carried out to demonstrate the effectiveness of the work.
\end{abstract}
\section{Introduction}
With the continued expansion of the robotics industry, the scope of interactions between robots and humans in everyday life is poised to increase, thereby placing higher demands on the intelligent evolution of robots. Conventional methodologies for robot learning detect the environment through sensors, coupled with extensive computational processes executed within simulated environments, all in the pursuit of developing logical motion planning strategies for robots during task execution \cite{Zhou_2022_JIM}. This approach, however, requires substantial time and has high requirements on hardware performance. In stark contrast, human execution of analogous tasks is simple and intuitive. Therefore, one promising way to enhance the robot intelligence involves learning from human demonstration, wherein humans assume the role of instructors. Within this framework, robots imitate and learn from demonstrations (LfD), thereby elevating their behavioral dexterity.

One work of LfD involves the acquisition of human-guided instructional data. Conventional approaches to data collection employ mechanical sensors to gather information from human actions. In \cite{Chen_2021_TNSRE}, Chen et al. utilized motion capture markers and an Inertial Measurement Unit (IMU) to capture the foot movement. In parallel with advancements in computer vision technologies, the utilization of cameras has emerged as an alternative mechanism for capturing the human demonstration data. A notable advantage of employing cameras lies in obviating the necessity for individuals to sensors, thereby offering a more expeditious and streamlined alternative to conventional data collection methods. In \cite{Cai_2016_TRO}, Cai et al. used a single camera to track the position of human-driven objects, facilitating the subsequent emulation of these trajectories by robotic systems.

In recent years, a proliferation of camera-based skeletal detection tools has emerged, among which OpenPose, introduced by Cao et al. in 2017 \cite{Cao_2017_CVPR}. It enables the real-time extraction of human skeletal structures from webcam feeds and is amenable to multi-person scenarios, although it demands relatively high hardware requirements. In \cite{Fang_2023_TPAMI}, Fang et al. localized whole-body keypoints accurately and tracked humans simultaneously with OpenPose. However, for fine tasks, it is insufficient to track the Cartesian coordinates of the human body; it also requires the orientation of parts of the human body, such as the hands. For example, in \cite{Li_2022_TCYB}, Li et al. extracted factors from hand heatmaps to estimate hand poses and teleoperate a dual-arm system. MediaPipe is another vision-based tool for human skeletal keypoint extraction \cite{Lugaresi_2019_CVPR}. In comparison to OpenPose, MediaPipe holds the advantage of accurately and efficiently capturing two-dimensional (2D) key points of the human hand, thus facilitating precise hand gesture acquisition. In \cite{Chen_2023_CSME}, Chen et al. utilized two cameras to capture 2D points of the hand and generate the three-dimensional (3D) coordinates to obtain trajectories of the human hand.
\begin{figure*}[t]
    \centering
    \includegraphics[scale=0.7]{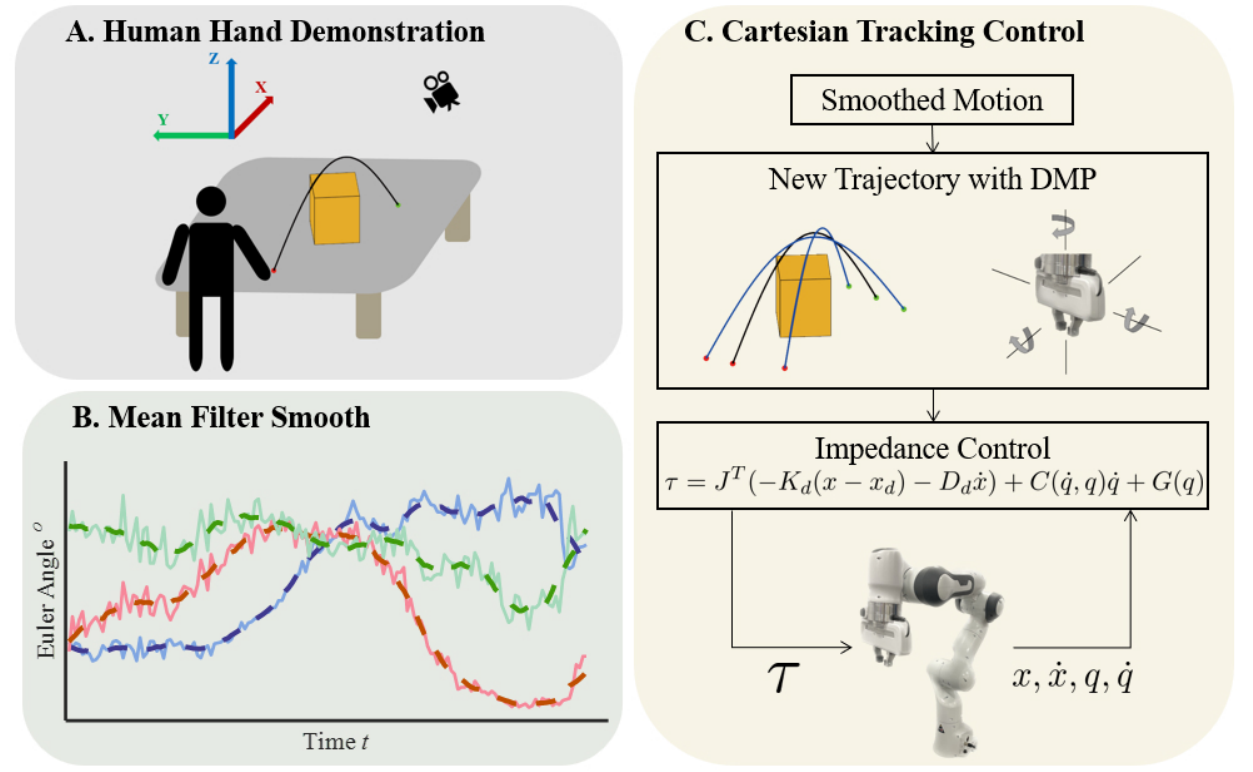}
    \caption{The Schematic Diagram of the Proposed Work}
    \label{Diagram}
\end{figure*}

Subsequent to the reception of human-guided instructions, robots have to learn from human actions. Behavioral cloning is a method to duplicate the human behavior by collecting the demonstration data, and the input-output mapping is simply supervised by learning methods \cite{Ly_2021_TIV}. In comparison to behavioral cloning, reinforcement learning (RL) offers a more flexible and adaptive approach to learning. In \cite{Liu_2022_RAL}, an inverse RL infers the latent reward function of a task from expert demonstrations to better understand the task structure and generalize to novel situations. While the inverse RL method requires a more substantial volume of data and computational resources, Dynamic Movement Primitives (DMP) emerges as a notable methodology for robotic motion generation and control with a single demonstration \cite{Ijspeert_2002_ICRA}. DMP aims to simulate the dynamic properties and flexibility exhibited by human when performing motion. This enables the DMP to generate suitable motions when encountering new situations, and allows for fine-tuning and adaptation while following prescribed trajectories. In \cite{Han_2023_TCYB}, the integration of DMP with adaptive admittance control enables the path planning and force control on curved surfaces.

Several works have integrated DMP with vision sensors to control the manipulator. In \cite{Chen_2023_IFAC}, Chen et al. utilized You Only Look Once (YOLO) to train and detect hand shapes with two webcams, enabling the robot to pick up an object and place it in a human hand. DMP was applied to set trajectories of the end-effector, learned from dragging movements. Similarly, Cai et al. detected multiple hand demonstrations using a depth camera and OpenPose to obtain a comprehensive translational trajectory and predict the endpoint by  DMP \cite{Cai_2023_IFAC}. However, these works did not explicitly consider the hand's quaternions in motion planning. To the best of the author's knowledge, there has been no application of DMP with both translational and rotational demonstrations captured by cameras. Incorporating quaternions in motion planning adds a layer of dexterity, and exploring this aspect could be a potential avenue for future research in enhancing manipulator control.

The proposed framework is shown as in Fig. \ref{Diagram}. In this paper, a depth camera is employed with MediaPipe applied in generating 2D images which are then combined with the depth data to capture the 3D coordinates of the whole human hand. This enables the recording of the trajectory, orientation, and grasping of its movements. To mitigate the impact of the inevitable minor tremors in human motion, the acquired human demonstration data undergoes a pre-processing phase with proposed method to calculate the orientation of the hand and finger motions, involving the application of a mean filter. Following pre-processing, a modified DMP is proposed to learn the coordinate trajectory of the wrist. The new trajectories with novel start and end points are applied to the execution of the pick-and-place task. This task entails the precise manipulation of objects, such as the experimental demonstration including picking up object at a specified angle, avoiding obstacles, and the ultimate placement of the object within an inclined receptacle. The  proposed framework offers a novel effective approach to many dexterous manipulation tasks. 
\section{System Description}
\subsection{Robotic Manipulator}
The equipment used in the experiment is the 7-degree-of-freedom (7-DOF) Franka Emika robot, which can perform complex tasks with dexterity. The joints provide a signal transmission frequency of $1$ $kHz$ to ensure smooth data process. The dynamics of the Franka Emika robot manipulator in joint space is presented as in Eq.~\eqref{dynamic_js}:
\begin{equation}
    \mathcal{M}(\bm{q})\bm{\ddot{q}}+\mathcal{C}(\bm{\dot{q}},\bm{q})\bm{\dot{q}}+\mathcal{G}(\bm{q})=\bm{\tau},\label{dynamic_js}
\end{equation}
where $\mathcal{M}(\bm{q})\in\mathbb{R}^{7\times7}$ represents the inertial matrix, $\mathcal{C}(\bm{\dot{q}}, \bm{q})\in\mathbb{R}^{7\times7}$ represents the Coriolis and centripetal matrix, $\mathcal{G}(\bm{q})\in\mathbb{R}^{7}$ is the gravity vector and \bm{$\tau$}$\in\mathbb{R}^{7}$ is the torque input vector. \bm{$q$}, \bm{$\dot{q}$}, \bm{$\ddot{q}$}$ \in \mathbb{R}^{7}$ are the joint angle, velocity, and acceleration vectors.

To accomplish trajectory tracking in the experiment, the dynamic equation can be transformed to that in Cartesian space. The end-effector pose is denoted as 
$\bm{x} = \begin{bmatrix}
\bm{p}, & \bm{\xi}
\end{bmatrix}^T\in\mathbb{R}^{7}$ where $\bm{p}=
\begin{bmatrix}
    x, & y, & z
\end{bmatrix}\in\mathbb{R}^{3}$ is the position in Cartesian space and $\bm{\xi} = 
\begin{bmatrix}
    q_{w}, & q_{x}, & q_{y}, & q_{z}
\end{bmatrix}\in\mathbb{R}^4$ 
is the quaternion, where $q_{w}$ denotes the real part, and $q_{x}, q_{y}, q_{z}$ denote the imaginary part. The torque input vector $\bm{\tau}$ is transformed to the force control input $\bm{u}\in\mathbb{R}^{6}$. The transformation from the joint space to the Cartesian space is shown in Eq.~\eqref{js_to_cs}.
\begin{equation}
    \left[\begin{matrix}
        \bm{\dot{x}}\\
        \bm{\ddot{x}}
    \end{matrix}\right]
    =
    \left[\begin{matrix}
        J(\bm{q})\bm{\dot{q}}\\
        J(\bm{q})\bm{\ddot{q}}+\dot{J}(\bm{q})\bm{\dot{q}}
    \end{matrix}\right],\label{js_to_cs}
\end{equation}
where $J(\bm{q})\in\mathbb{R}^{6\times7}$ is the Jacobian matrix, $\dot{\bm{x}} = 
\begin{bmatrix}
    \dot{\bm{p}}, & \bm{\omega}
\end{bmatrix}^T \in \mathbb{R}^{6}, \ddot{\bm{x}} =
\begin{bmatrix}
    \ddot{\bm{p}}, & \dot{\bm{\omega}}
\end{bmatrix}^T \in \mathbb{R}^{6}$, where $\bm{\omega}, \ \dot{\bm{\omega}} \in \mathbb{R}^3$ are the angular velocity and acceleration of the end-effector, respectively. 

With Eq.~\eqref{dynamic_js} and Eq.~\eqref{js_to_cs}, the dynamic equation of the manipulator in Cartesian space can be presented in Eq.~\eqref{dynamic_cs}.
\begin{equation}
    \mathcal{\bar{M}}(\bm{q})\bm{\ddot{x}}+\mathcal{\bar{C}}(\bm{\dot{q}},\bm{q})\bm{\dot{x}}+\mathcal{\bar{G}}(\bm{q})=\bm{u},\label{dynamic_cs}
\end{equation}
where
    \begin{align*}
        \bar{\mathcal{M}}(\bm{q}) &= J(\bm{q})^{-T} \mathcal{M}(\bm{q}) J(\bm{q})^{-1}, \\
        \bar{\mathcal{C}}(\dot{\bm{q}},\bm{q}) &= J(\bm{q})^{-T} ( \mathcal{C}(\dot{\bm{q}},\bm{q}) 
         - \mathcal{M}(\bm{q}) J(\bm{q})^{-1} \dot{J}(\bm{q}) ) J(\bm{q})^{-1},\\
        \bar{\mathcal{G}}(\bm{q}) &= J(\bm{q})^{-T} \mathcal{G}(\bm{q}),\ \ \text{and} \ \ 
        \bm{u} = J(\bm{q})\bm{\tau}.
    \end{align*}
\subsection{Depth Camera}
The depth camera employed in this paper is the RealSense D435i, a product developed by Intel. This device is equipped with a dual camera system that captures visible light images and depth information. It relies on a laser projector and an infrared camera to measure the distance between objects and the camera, resulting in high quality depth images.
\section{Methodology}
\subsection{Human Hand Demonstration}
\subsubsection{3D Coordinate Generation of a Hand}
\begin{figure}[t]
    \centering
    \subfloat[RGB Image with MediaPipe]{
        \includegraphics[scale=0.5]{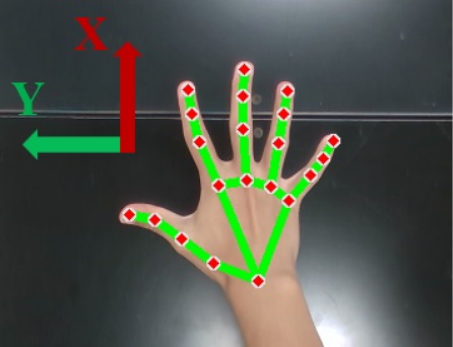}}
    \subfloat[Depth Image]{
        \includegraphics[scale=0.25]{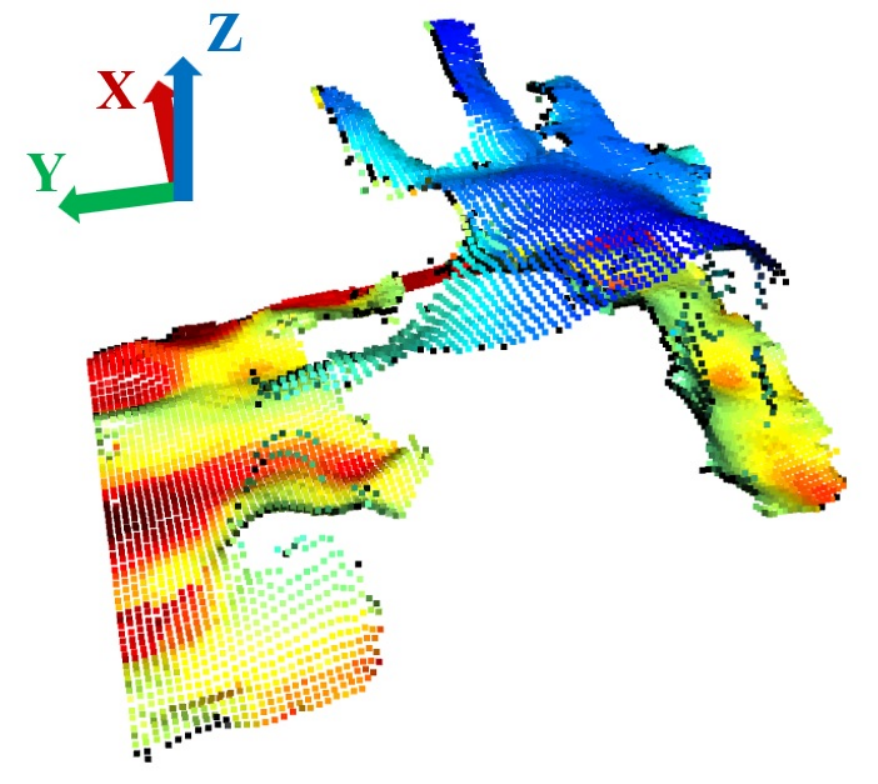}}
    \\
    \subfloat[3D Hand]{
        \includegraphics[scale=0.2]{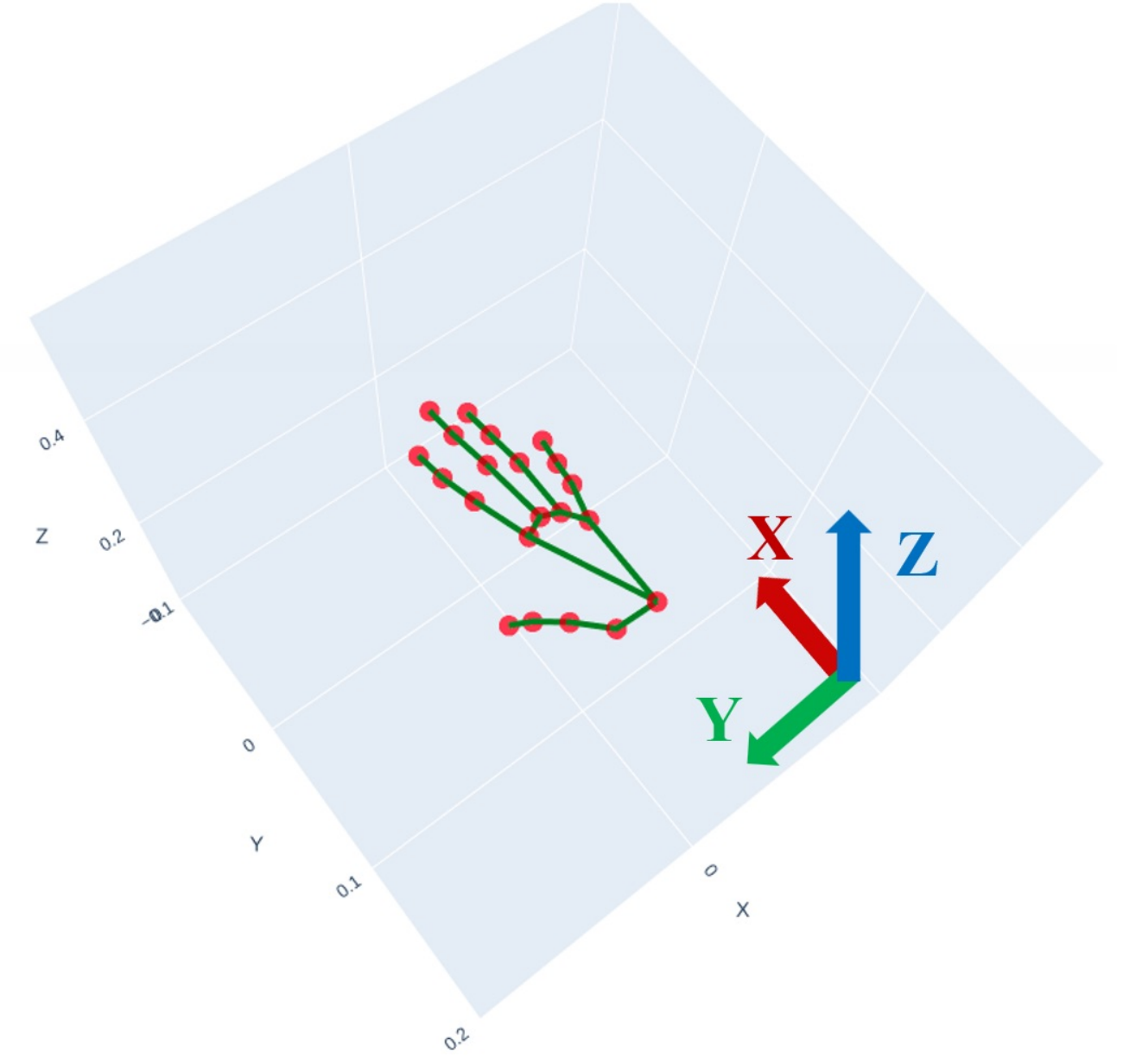}}
   \caption{The Proposed 3D Hand Coordinate Generation}
   \label{3D} 
\end{figure}

A depth camera can simultaneously capture Red-Green-Blue (RGB) images as well as the depth image while ensuring their alignment. In this paper, a consistent resolution of $640\times480$ was uniformly established for both the RGB and depth image. This standardization facilitates an accurate correspondence between the data points within these two distinct graphical representations. As shown in Fig. \ref{3D}(a), the initial step entails the application of the MediaPipe's hand detection function to identify the hand's key points within the RGB image. Subsequently, the 2D pixel coordinates of these 21 key points are obtained. Corresponding the index of these pixels to the depth image, the depth of these pixels can be obtained in Fig.\ref{3D}(b). The pixel coordinates do not represent real-world coordinates and therefore, a coordinate transformation from the pixel coordinates, $x_p$, $y_p$ to real-world spatial coordinates, $\bm{x}_h = \begin{bmatrix}
    \bm{p}_h^T, & \bm{\xi}_h^T
\end{bmatrix}^T$, is required.

In real-world spatial coordinates, an actual distance $l$ can be calculated by Eq.~\eqref{actual_distance}.
\begin{equation}
    l=d\tan(\frac{n}{N}\theta),\label{actual_distance}
\end{equation}
where $d$ is the depth, $n$ is the pixel distance, $N$ is the pixel value of width or height of the camera image, and $\theta$ is the view angle of the camera. Term $\frac{1}{N}\theta$ is one pixel's angle in the figure, and $d\tan(\frac{1}{N}\theta)$ is the real length of one pixel, so the actual distance $l$ with $n$ pixels can be concluded as in Eq.~\eqref{actual_distance}.

Using Eq.~\eqref{actual_distance}, we can get the 3D coordinates as in  Eq.~\eqref{xyz}.
\begin{equation}
    \bm{p}_{h}
    =
    \left[\begin{matrix}
    x_{h},y_{h},z_{h}
    \end{matrix}\right]^{T}
    =
    \left[\begin{matrix}
        d\tan(\frac{(x_{p}-\frac{X}{2})}{X}\theta_x)\\
        d\tan(\frac{(y_{p}-\frac{Y}{2})}{Y}\theta_y)\\
        H-d
    \end{matrix}\right],\label{xyz}
\end{equation}
where $H$ is the height of the camera, $\theta_x$ and $\theta_y$ are the view angles of the camera, $X$ and $Y$ are the resolution. $\theta_x,~\theta_y,~X,~Y$ are constant parameters related to the camera. In this paper, $~H=1.0 m,~X=640,~Y=480,~\theta_x=69^o,~\theta_y=42^o$. The final 3D hand is shown in Fig. \ref{3D}(c).

\subsubsection{Orientation}
In addition to the precise control of the 3D coordinates of the end-effector, equal significance is attributed to managing the orientation of the end-effector and the grasping of the gripper. These key parameters can be calculated and corresponded through the 3D coordinates of the thumb, index finger, and wrist. To represent the orientation of the end-effector, the Euler angles of yaw, pitch and roll orientations are as in Eq. \eqref{ypr}.
\begin{equation}
    \left[\begin{matrix}
        \psi_h\\
        \theta_h\\
        \phi_h
    \end{matrix}\right]
    =
    \left[\begin{matrix}
        \arctan(\frac{x_{i}-x_{t}}{y_{t}-y_{i}})\\
        \arctan(\frac{z_{w}-z_{t}}{x_{t}-x_{w}})\\
        \arctan(\frac{z_{i}-z_{t}}{y_{t}-y_{i}})
    \end{matrix}\right],\label{ypr}
\end{equation}
where the yaw angle $\psi_{h}$ is the rotation about the $Z$-axis, the pitch angle $\theta_{h}$ is the rotation about the $Y$-axis and the roll angle $\phi_{h}$ is the rotation about the $X$-axis. $(x_{i}, y_{i}, z_{i})$ denote the positions of index finger, $(x_{t},y_{t},z_{t})$ and $(x_{w},y_{w},z_{w})$ denote the positions of thumb and wrist, respectively.

While Euler angles offer a straightforward and intuitive method for the orientation representation, the Franka Emika Panda robot uses quaternions as its chosen representation for orientation, so the conversion from Euler angles to quaternions is needed. Prior to executing this transformation, it is imperative to understand the quaternion multiplication operation. Assume $\bm{\xi_{1}}$ and $\bm{\xi_{2}}$ are quaternions, which can be represented by Eq. \eqref{quaternions}.
\begin{equation}
    \left\{
        \begin{aligned}
            \bm{\xi_{1}} &= w_{1}+\bm{v_{1}}=w_{1}+x_{1}\bm{i}+y_{1}\bm{j}+z_{1}\bm{k},\\
            \bm{\xi_{2}} &= w_{2}+\bm{v_{2}}=w_{2}+x_{2}\bm{i}+y_{2}\bm{j}+z_{2}\bm{k}.
        \end{aligned}
    \right.\label{quaternions}
\end{equation}

Then the multiplication of $\bm{\xi_{1}}$ and $\bm{\xi_{2}}$ can be obtained as 
%
\begin{equation}
  \begin{aligned}
    \bm{\xi_{1}\xi_{2}} &= w_{1}w_{2}-\bm{v_{1}\cdot v_{2}}+w_{2}\bm{v_{1}}+w_{1}\bm{v_{2}}+\bm{v_{1}\times v_{2}}\\
            &=     
                \left[\begin{matrix}
                    x_{1}w_{2}+w_{1}x_{2}+y_{1}z_{2}-z_{1}y_{2}\\
                    y_{1}w_{2}+w_{1}y_{2}+z_{1}x_{2}-x_{1}z_{2}\\
                    z_{1}w_{2}+w_{1}z_{2}+x_{1}y_{2}-y_{1}x_{2}\\
                    w_{1}w_{2}-x_{1}x_{2}-y_{1}y_{2}-z_{1}z_{2}
                \end{matrix}\right].
  \end{aligned}\nonumber
\end{equation}

Through the multiplication of three axes, the transformation equation between quaternions and Euler angles is as: 
\begin{equation}
  \begin{aligned}
        \bm{\xi}_{h}
            &= \bm{\xi}_{x}(\phi_{h})\bm{\xi}_{y}(\theta_{h})\bm{\xi}_{z}(\psi_{h})\\
            &=     
                \left[\begin{matrix}
                    C_{\frac{\phi_{h}}{2}}\\
                    S_{\frac{\phi_{h}}{2}}\\
                    0\\
                    0
                \end{matrix}\right]
                \left[\begin{matrix}
                    C_{\frac{\theta_{h}}{2}}\\
                    0\\
                    S_{\frac{\theta_{h}}{2}}\\
                    0 
                \end{matrix}\right]
                \left[\begin{matrix}
                    C_{\frac{\psi_{h}}{2}}\\
                    0\\
                    0\\
                    S_{\frac{\psi_{h}}{2}}
                \end{matrix}\right]\\
            &=     
                \left[\begin{matrix}
                    C_{\frac{\phi_{h}}{2}}C_{\frac{\theta_{h}}{2}}C_{\frac{\psi_{h}}{2}}+S_{\frac{\phi_{h}}{2}}S_{\frac{\theta_{h}}{2}}S_{\frac{\psi_{h}}{2}}\\
                    S_{\frac{\phi_{h}}{2}}C_{\frac{\theta_{h}}{2}}C_{\frac{\psi_{h}}{2}}-C_{\frac{\phi_{h}}{2}}S_{\frac{\theta_{h}}{2}}S_{\frac{\psi_{h}}{2}}\\
                    C_{\frac{\phi_{h}}{2}}S_{\frac{\theta_{h}}{2}}C_{\frac{\psi_{h}}{2}}+S_{\frac{\phi_{h}}{2}}C_{\frac{\theta_{h}}{2}}S_{\frac{\psi_{h}}{2}}\\
                    C_{\frac{\phi_{h}}{2}}C_{\frac{\theta_{h}}{2}}S_{\frac{\psi_{h}}{2}}-S_{\frac{\phi_{h}}{2}}S_{\frac{\theta_{h}}{2}}C_{\frac{\psi_{h}}{2}}
                \end{matrix}\right],\nonumber
  \end{aligned}
\end{equation}
where $C_{x} = \cos(x)$ and $S_{x} = \sin(x)$ with $x = \frac{\theta_{h}}{2}, \frac{\psi_{h}}{2}, \frac{\phi_{h}}{2}$.
\begin{figure}[t]
    \centering
    \subfloat[]{
	   \includegraphics[scale=0.15]{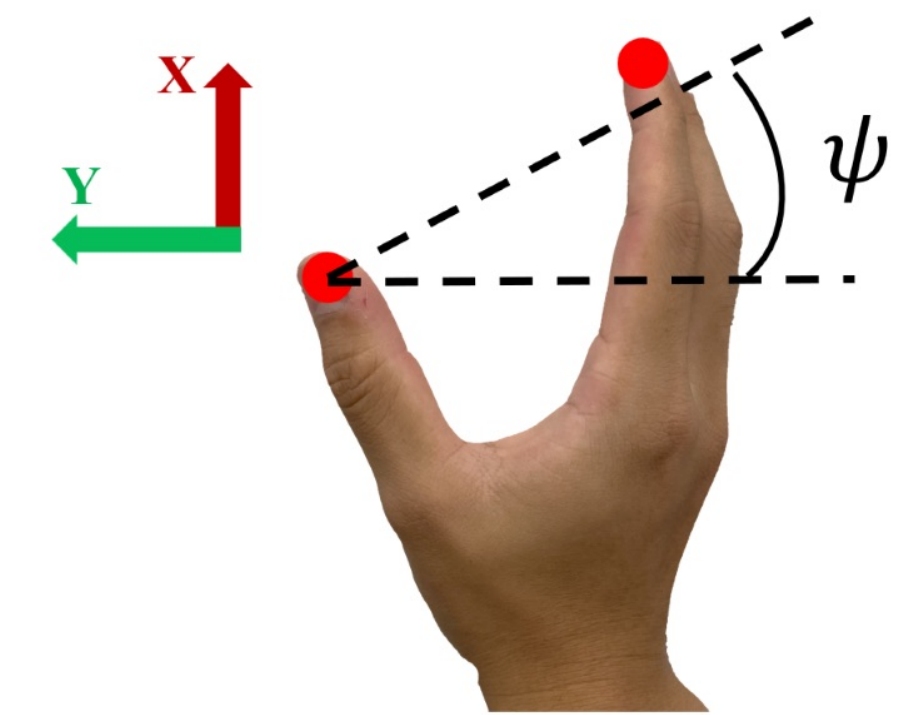}
    }
    \subfloat[]{
	   \includegraphics[scale=0.15]{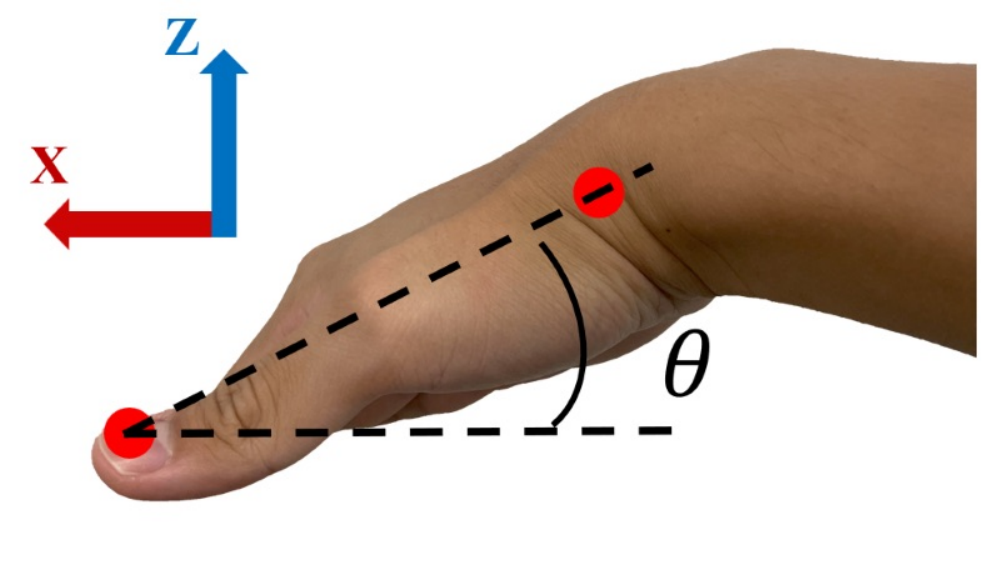}
    }
    \subfloat[]{
	   \includegraphics[scale=0.15]{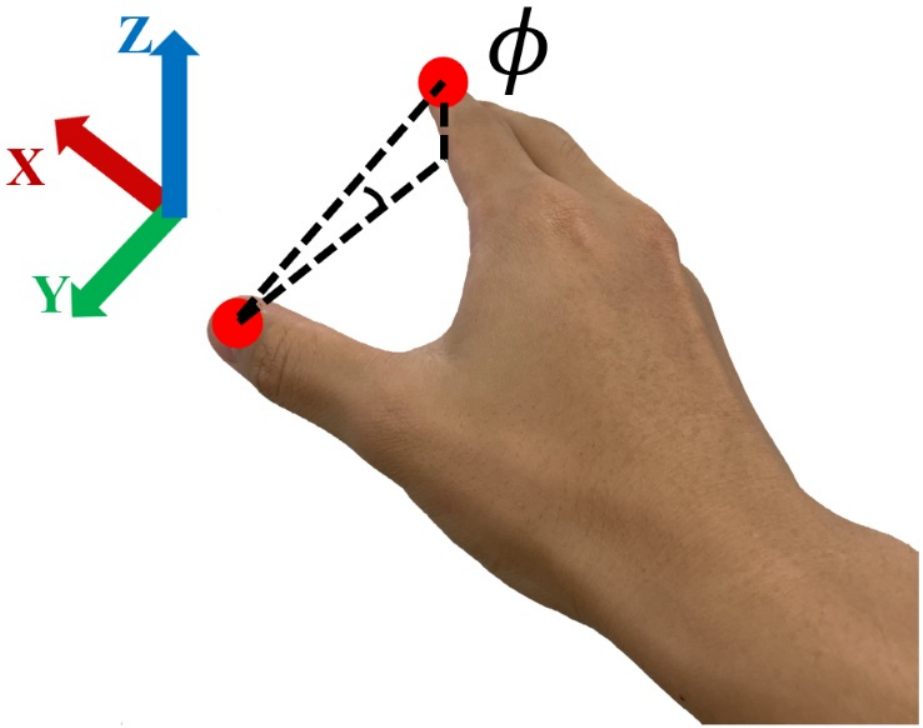}
    }
    \\
    \subfloat[]{
        \includegraphics[scale=0.15]{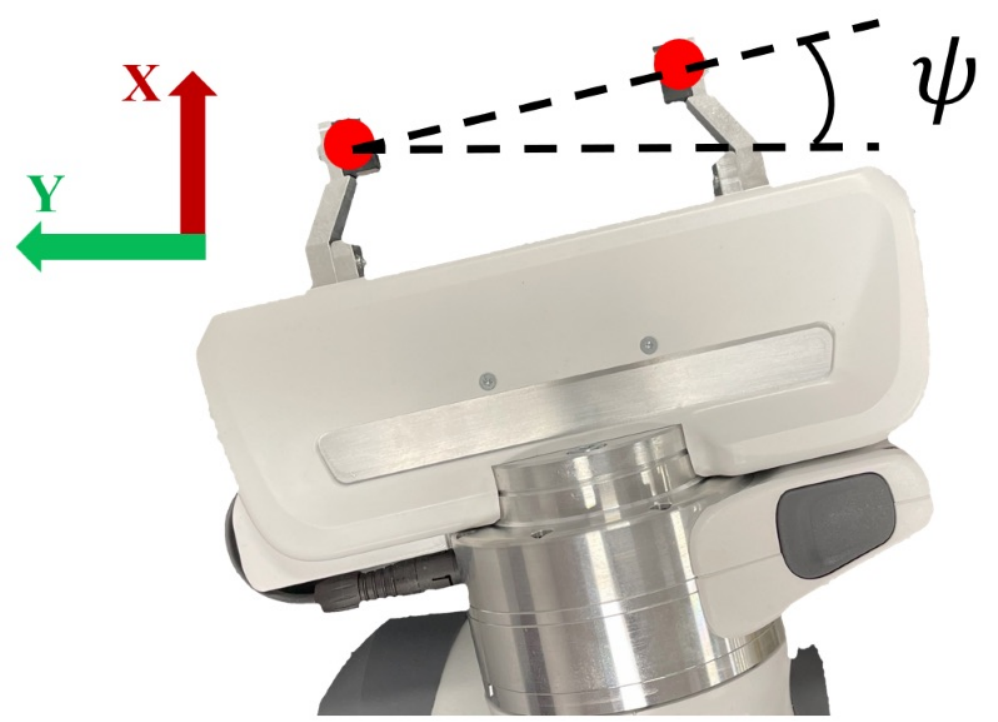}
    }
    \subfloat[]{
        \includegraphics[scale=0.15]{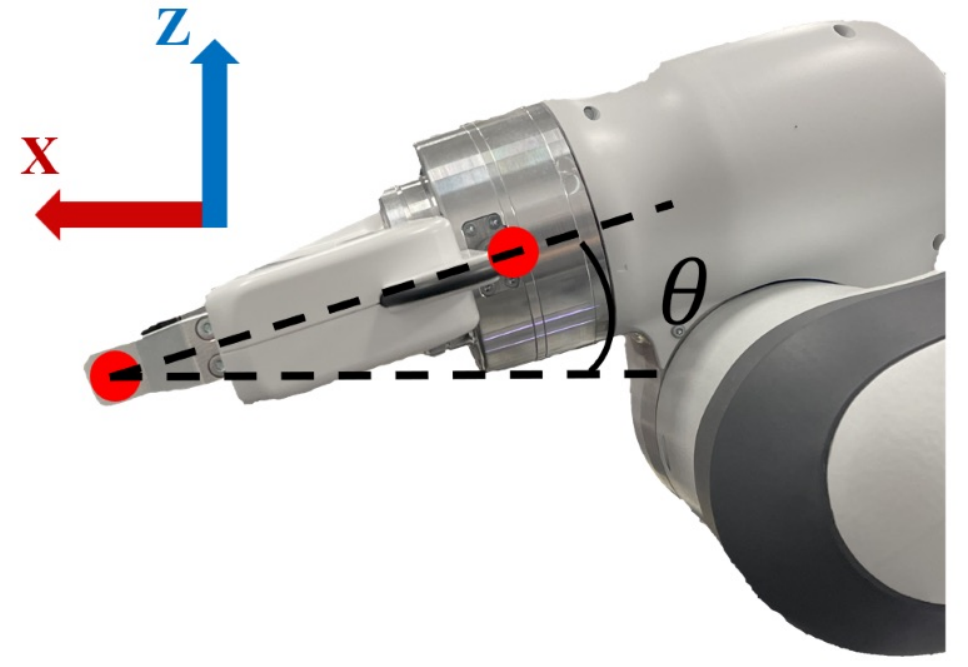}
    }
    \subfloat[]{
        \includegraphics[scale=0.15]{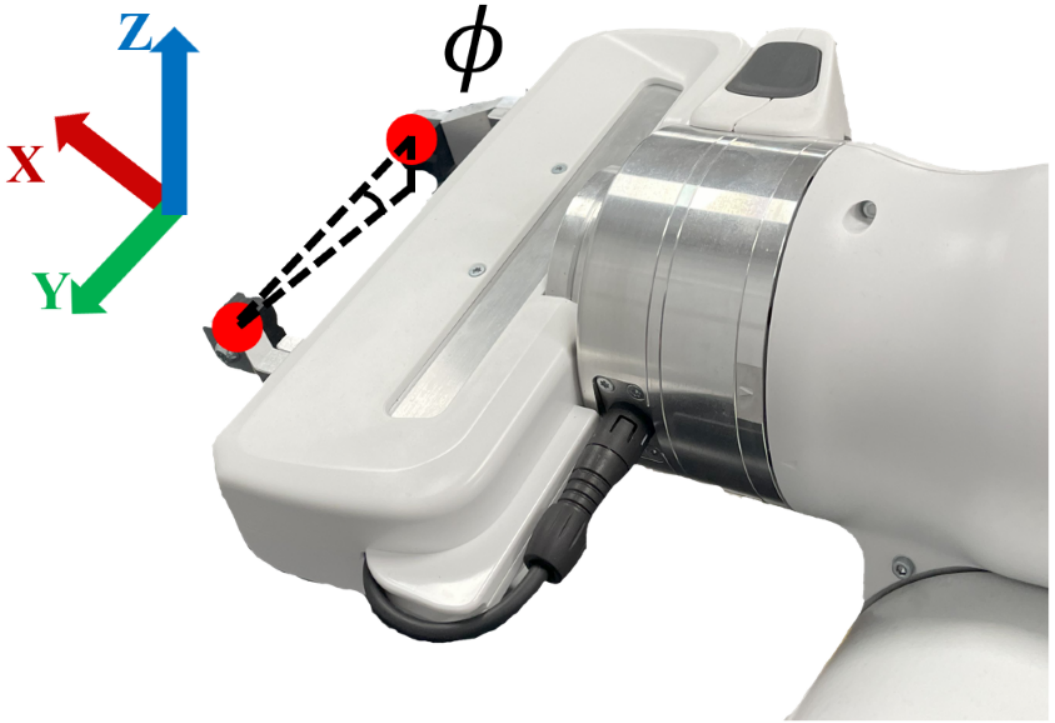}
    }
    \caption{Euler Angles Generation. (a)-(c) are the yaw, pitch, and roll of hand and (d)-(f) are the yaw, pitch, and roll of gripper.}
    \label{fig3}
\end{figure}

Given that the default configuration of the robot's end-effector is perpendicular to the ground,  while the default hand posture in the human demonstration aligns parallel to the ground, an essential adjustment is mandated. We rotate the end-effector $90^{\circ}$ around the $Y$-axis, so the desired quaternion $\bm{\xi}_d$ should be calculated as the demonstration quaternion $\bm{\xi}_{h}$ multiplying the quaternion rotated $90^{\circ}$ around the $Y$-axis: 
\begin{equation}
    \bm{\xi}_{d}= \bm{\xi}_{h}\bm{\xi}_{y}(\frac{\pi}{2}).\nonumber\label{final_quaternion}
\end{equation}
\subsubsection{Grasping}
For grasping, the distance between the thumb and the index finger $d_{ti}$ can be calculated as Eq. \eqref{distance_ti}. If the distance is smaller than the threshold, robot will consider it as a grasping motion learned and the gripper will then close and grasp. In this paper, the threshold is set to $0.10 m$ which can be changed depending on the tasks. 
\begin{equation}
    d_{ti}=\sqrt{(x_{t}-x_{i})^2+(y_{t}-y_{i})^2+(z_{t}-z_{i})^2}.\label{distance_ti}
\end{equation}
\subsubsection{Data Pre-processing}
After obtaining the motion trajectory and posture from the human demonstration, a mean filter is applied to smooth the raw data. The mean filter requires a one-dimensional vector of length $N$, denoted as $\bm{x_h}$, and the output vector $\bm{x^{*}_{h}}$ after applying an average smoothing filter with a window size of $k$ given by Eq. \eqref{mean_filter}.
\begin{equation}
    x^{*}_{hi} = \frac{1}{k} \sum_{j=i-\frac{k-1}{2}}^{i+\frac{k-1}{2}} x_{hj},\label{mean_filter}
\end{equation}
where $x^{*}_{hi}$ represents the $i$th element in the output vector, $x_{hj}$ corresponds to the $j$th element in the input vector and $i$ ranges from $\frac{k-1}{2}$ to $N-\frac{k-1}{2}$.
\subsection{Motion Planning}

\subsubsection{Original Dynamic Movement Primitives}

In \cite{Ijspeert_2002_ICRA}, it is proposed that complex actions are composed of a set of primitive actions that are executed sequentially or in parallel, and DMP is the mathematical formalization of these primitive actions. In fact, DMP serves as an approach to decompose complex motion into a set of basic motion primitives. Each motion primitive is characterized as a nonlinear system whose dynamic properties are influenced by a guided trajectory, such that the primitives can be reutilized and adapted across various settings. 

At its core, the system model of DMP is characterized by the fusion of a Proportional-Derivative (PD) controller with the inclusion of the control term $f$, which, notably, is a nonlinear function. In this way, the system can not only converge to the goal point, but also $f$ allows the motion process to emulate the original trajectory. The dynamic system can be presented as Eq. \eqref{DMP_system}.
\begin{equation}
    \left\{
        \begin{aligned}
            \delta\dot{\hat{\bm{v}}}_h &= K(\bm{g}_h-\bm{\hat{x}}_h)-D \bm{\hat{v}}_h+(\bm{g}_h-\bm{\hat{x}}_{h_0}) \bm{f_h},\\
            \delta\bm{\dot{\hat{x}}}_h &= \bm{\hat{v}}_h,
        \end{aligned}\label{DMP_system}
    \right.
\end{equation}
where $\bm{\hat{x}}_h$ and $\bm{\hat{v}}_h$ are the position and velocity of the system, $\bm{\hat{x}}_{h_0}$ and $\bm{g}_{h}$ are the start and goal points of the trajectory, $\delta$ is the time duration, $K$ is the spring stiffness, and $D$ is the damper damping.

In order to generate $f$, it is imperative to first acquire ${\bm{f_{h}}}_{target}$, which can be represented by the demonstration trajectory as Eq. \eqref{f_target}.
\begin{equation}
    {\bm{f_{h}}}_{target}=\frac{\delta \ddot{\bm{x}}^*_{h}-K(\bm{g}_h-\bm{x}^*_h)+D\delta \dot{\bm{x}}^*_h}{\bm{g}_h-\bm{\hat{x}}_{h_0}},\label{f_target}
\end{equation}
where $\bm{x}^{*}_{h}$, $\dot{\bm{x}}^{*}_{h}$, $\ddot{\bm{x}}^{*}_{h}$ are the position, velocity and acceleration of the pre-processed demonstration trajectory. $\bm{f_{h}}$ is the nonlinear function used to generate arbitrary complex movement, so the work in \cite{Ijspeert_2002_ICRA} used Gaussian functions as the basis functions to represent $\bm{f_{h}}$. Assume that each element, $f_{h}$ , has its own set of parameters. The basis functions are:
\begin{equation}
   f_{h}=\frac{\sum_{i=1}^{N}\omega_{i}\Psi_{i}(s)}{\sum_{i=1}^{N}\Psi_{i}(s)}s,\nonumber\label{f}
\end{equation}
where 
    \begin{align*}
        \dot{s} = -\alpha_{s}s,\ \ \ 
        \Psi_{i}(s) = \exp(-h_{i}(s-c_{i})^2),
    \end{align*}
and $s > 0$ starts at one and gradually tends toward zero, thereby ensuring that $\bm{f_{h}}$ approaches zero when $\bm{\hat{x}}_{h}$ converges to $\bm{g_{h}}$. $\alpha_{s}$ is a constant value, $\Psi_{i}$ is the Gaussian function, where $c$ is the center, $h$ is the width, and $\omega$ is the adjustable weight. Each Gaussian function is endowed with a respective weight, and our goal is to find such a set of weights that minimizes the error between $\bm{f_{h}}$ and ${\bm{f_{h}}}_{target}$. Locally weighted regression is used to obtain $\omega_{i}$ as: 
\begin{equation}
   \omega_{i}=\frac{s_{x}^T\Gamma_{i}f_{h_{target}}}{s_{x}^T\Gamma_{i}s_{x}},\nonumber\label{omega}
\end{equation}
where
    \begin{align*}
        s_{x} =
            \left(\begin{matrix}
                s_{1}(g_{h}-\hat{x}_{h_{0}})\\
                s_{2}(g_{h}-\hat{x}_{h_{0}})\\
                \cdots\\
                s_{n}(g_{h}-\hat{x}_{h_{0}})
            \end{matrix}\right),
        \Gamma_{i} =
            \left(\begin{matrix}
                \Psi_{i}(1)&&&0\\
                &&\ddots&\\
                0&&&\Psi_{i}(n)
            \end{matrix}\right),
    \end{align*}
and $n$ is the number of sampling points.

\subsubsection{Modified Dynamic Movement Primitives}
In Eq. \eqref{DMP_system}, $(\bm{g}_h-\bm{\hat{x}}_{h_0}) \bm{f_h}$ poses a potential issue when the starting point of the demonstration closely approximates the target position. In such cases, the term $(\bm{g}_h-\bm{\hat{x}}_{h_0})$ approaches to zero, consequently driving the term $\bm{f_h}$ towards nullity. Additionally, the opposite signs of $\bm{g}_h$ and $\bm{\hat{x}}_{h_0}$ engenders a mirroring effect in the trajectory shape. A modified DMP, as shown in Eq. \eqref{Modified_DMP}, wherein the system separates $\bm{f_h}$ and $(\bm{g}_h-\bm{\hat{x}}_{h_0})$ so that $\bm{f_h}$ remains unaffected by $\bm{g}_h$ and $\bm{\hat{x}}_{h_0}$ \cite{Park_2008_ICHR}.
\begin{equation}
    \left\{
        \begin{aligned}
            \delta\dot{\hat{\bm{v}}}_h &= K(\bm{g}_h-\bm{\hat{x}}_h)-D \bm{\hat{v}}_h-(\bm{g}_h-\bm{\hat{x}}_{h_0}) \bm{\hat{x}}_h + \bm{f_h},\\
            \delta\bm{\dot{\hat{x}}}_h &= \bm{\hat{v}}_h.
        \end{aligned}\label{Modified_DMP}
    \right.
\end{equation}

\subsection{Path Following Control}
In this paper, the manipulator is controlled by an impedance controller, which imparts a measure of flexibility through the modulation of stiffness in its movements. The principle of impedance control is to treat end-effector as a mass-spring-damper system. The torque $\bm{\tau}$ is designed in Cartesian space as
\begin{equation}
    \begin{aligned}
       \bm{\tau}=J^T(-K_d(\bm{x} - \bm{\hat{x}}_h) - D_d \dot{\bm{x}}) +C(\bm{\dot{q}},\bm{q})\bm{\dot{q}}+G(\bm{q}),\nonumber\\
    \end{aligned}
\end{equation}
where the gains $K_d$ and $D_d$ are the design parameters. 

\section{Performance Evaluation}
\subsection{3D Coordinate Accuracy}
This part of the experiment is dedicated to validating of the accuracy associated with the 3D coordinate generated by MediaPipe and Eq. \eqref{xyz}. The measured and calculated coordinates are shown in the Table I.
\begin{table}[htbp]
	\caption{Measurement of Position}
	\begin{center}
		\begin{tabular}{cccc}
                \hline
                Point & Measured (cm) & Calculated (cm) & Absolute Error (cm)\\
                \hline
                1 & (2.0, 8.0, 9.0) & (2.4, 8.0, 7.5) & (0.4, 0.0, 1.5)\\
                2 & (-5.0, 0.0, 9.0) & (-6.3, 0.7, 8.1) & (1.3, 0.7, 0.9)\\
                3 & (-6.0, -9.0, 34.5) & (-7.5, -8.1, 34.2) & (1.5, 0.9, 0.3)\\
                4 & (-19.0, 7.0, 26.5) & (-20.5, 7.4, 26.8) & (1.5, 0.4, 0.3)\\
                5 & (20.0, 10.0, 26.5) & (21.3, 10.1, 27.3) & (1.3, 0.1, 0.8)\\
                6 & (11.0, -10.0, 12.5) & (10.6, -8.6, 13.9) & (0.4, 1.4, 1.4)\\
                7 & (-14.0, -8.0, 12.5) & (-15.5, -6.8, 12.0) & (1.5, 1.2, 0.5)\\
                8 & (27.0, -14.0, 34.5) & (28.6, -12.2, 36.1) & (1.6, 1.8, 1.6)
                \end{tabular}
		\label{tab1}
	\end{center}
\end{table}

As shown in Table I, the maximum error observed along each axis remains confined within the threshold of $2$ $cm$. Some errors may be due to the measurement and the small shaking of the hand during demonstration. Noises are inevitable in the human demonstration data because human movements always have slight jitters. Hence a data smoothing approach is applied to the raw data.

\subsection{Data Pre-processing}
\begin{figure}[h]
    \centering
    \includegraphics[scale=0.55]{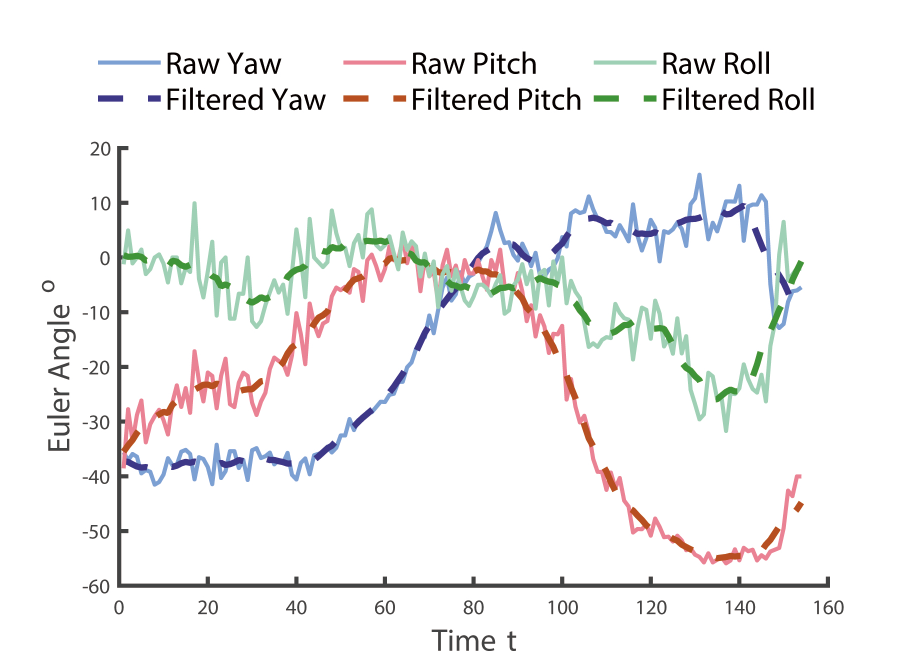}
    \caption{Smooth of Euler Angle}
    \label{fig:smooth}
\end{figure}

In light of the inherent noise present in the data collected from the human hand, a pre-processing step is employed. Specifically, we undertake data pre-processing through the application of a mean filter. Fig. \ref{fig:smooth} shows the comparison between raw and filtered Euler angles. The window size $k$ was tuned to 10.

\subsection{Dynamic Movement Primitives}

The execution of the human demonstration involves picking up a sponge from the workbench with 40-degree yaw, moving it over a cup and putting the sponge in a box sloped with 50-degree pitch. The trajectory and value of the task can be seen in Fig. \ref{fig5}. Fig. \ref{fig5} (d) and (e) show the Euler angles and distance $d_{ti}$ in the demonstration, which will be replicated by the end-effector. Then we employ modified DMP to learn the trajectory of $X$, $Y$, and $Z$, respectively, with three new starting points: $\begin{bmatrix}0.37,-0.34,0.22\end{bmatrix}$, $\begin{bmatrix}0.50,-0.25,0.21\end{bmatrix}$, $\begin{bmatrix}0.55,-0.34,0.28\end{bmatrix}$,  
and three new end points: $\begin{bmatrix}0.51,0.11,0.31\end{bmatrix}$, $\begin{bmatrix}0.50,0.19,0.30\end{bmatrix}$, $\begin{bmatrix}0.50,0.28,0.32\end{bmatrix}$. Three new trajectories are shown in Fig. \ref{fig5}(a)(b)(c)(f). New trajectories change the start and end points, but keep the shape, quaternion, and grasping motion. The video of the experiment can be seen in the ACM Lab YouTube channel: \url{https://www.youtube.com/watch?v=XP22mKGLvUI.}  
\begin{figure*}[t]
    \centering
    \subfloat[X]{
        \includegraphics[scale=0.37]{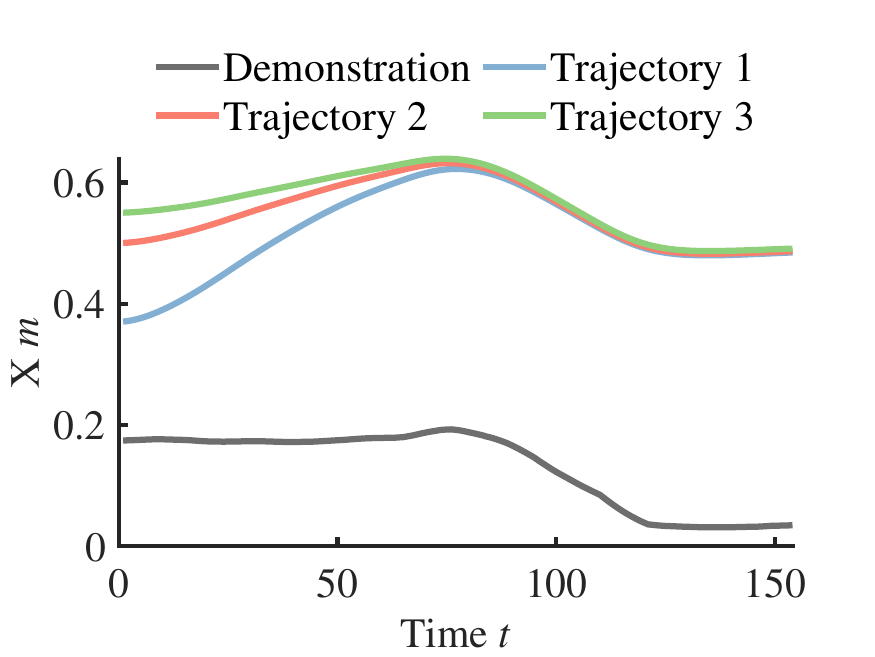}
    }
    \subfloat[Y]{
        \includegraphics[scale=0.37]{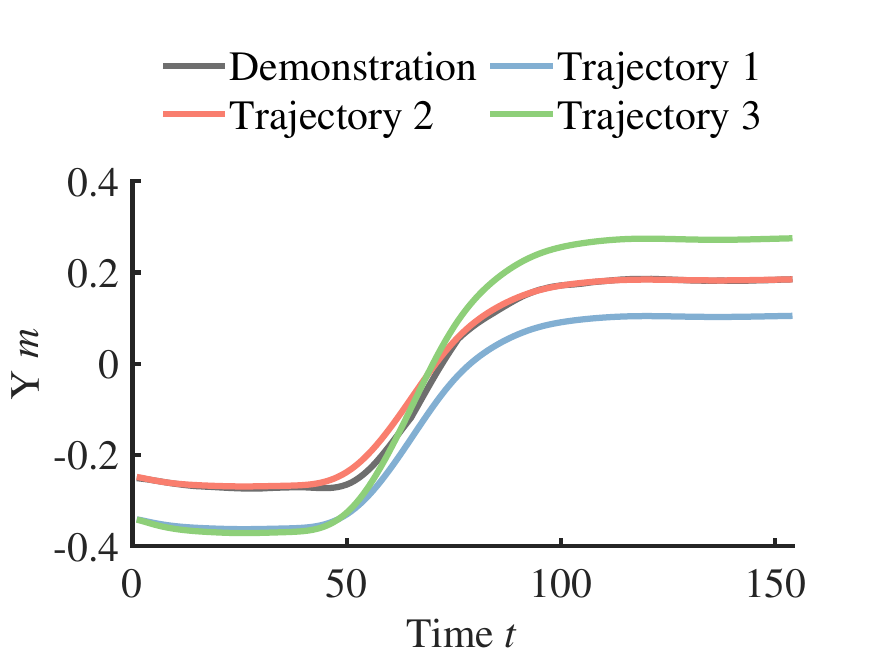}
    }
    \subfloat[Z]{
        \includegraphics[scale=0.37]{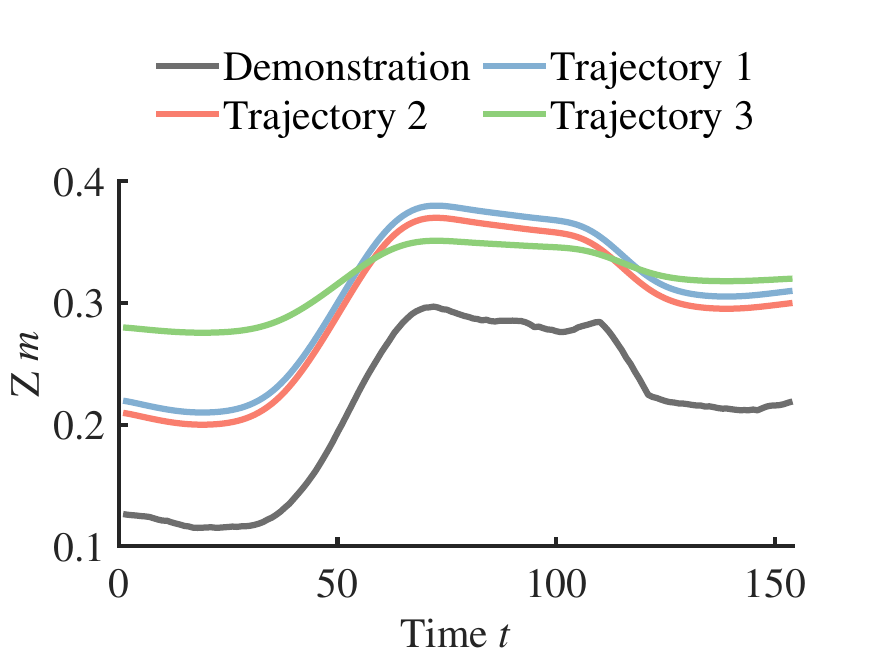}
    }
    \\
    \subfloat[Yaw, Pitch and Roll]{
        \includegraphics[scale=0.37]{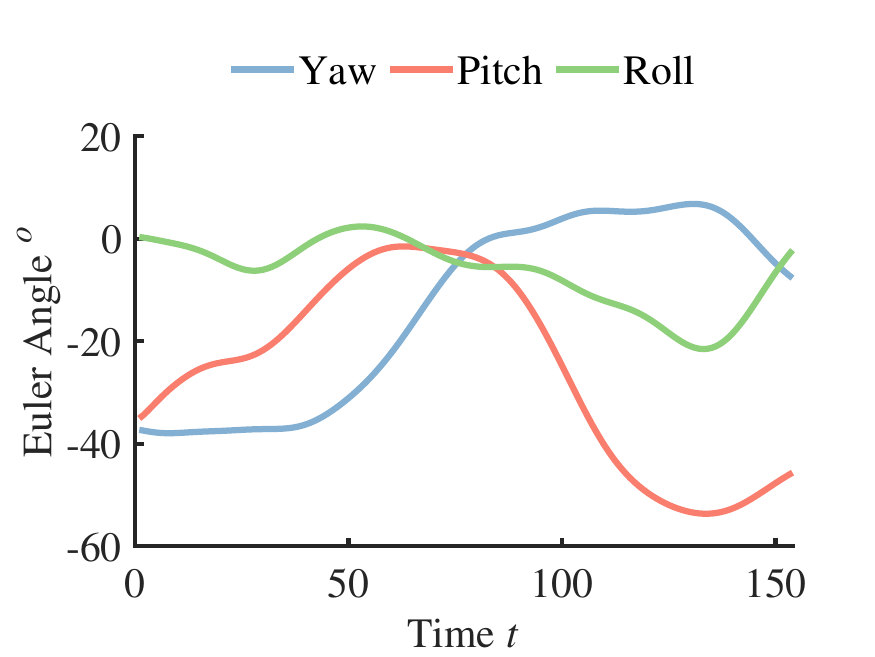}
    }
    \subfloat[Distance]{
        \includegraphics[scale=0.37]{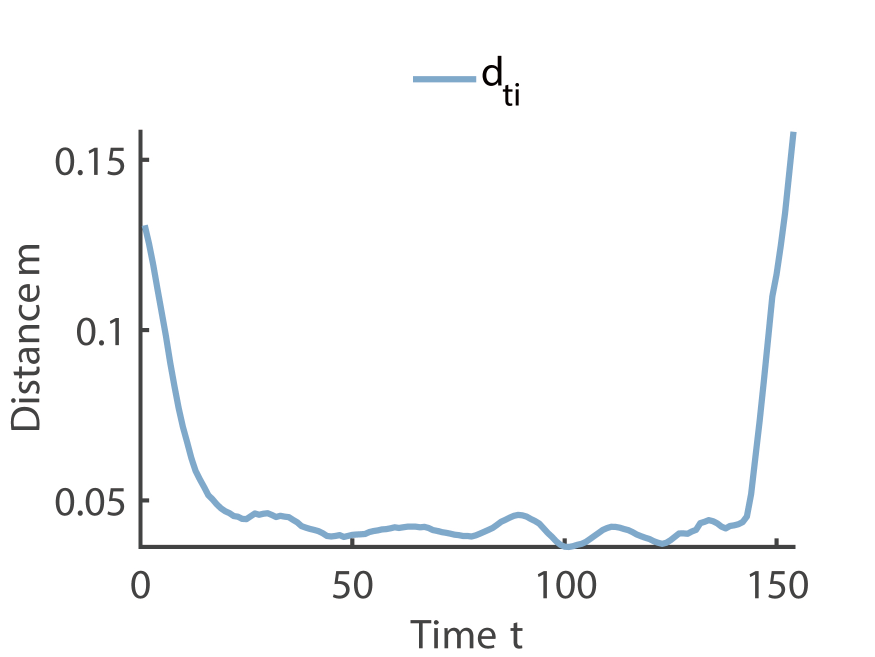}
    }
    \subfloat[3D Trajectory]{
        \includegraphics[scale=0.37]{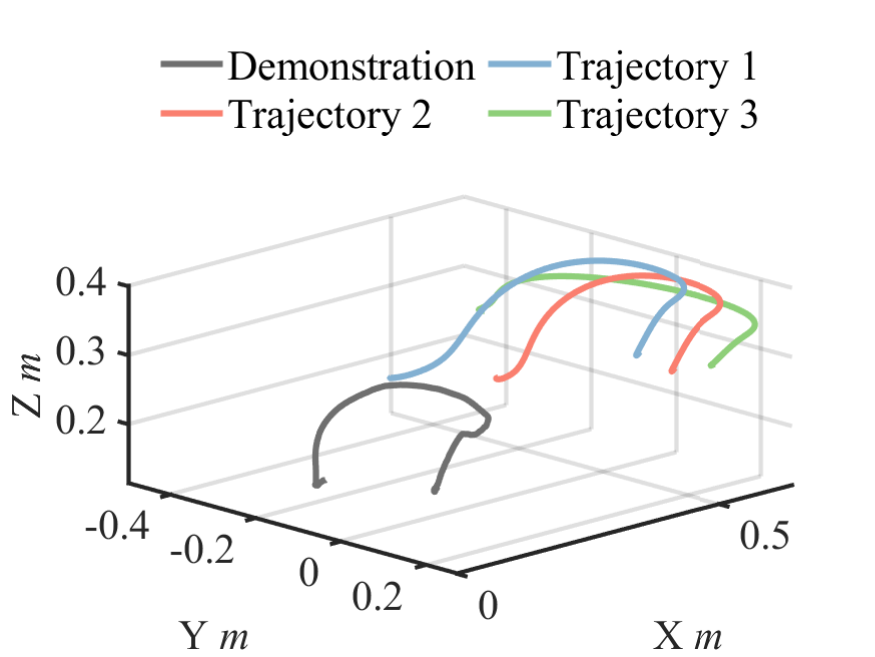}
    }
    \caption{Human demonstration and new trajectories generated by modified DMP.}
    \label{fig5}
\end{figure*}
%
\section{Conclusion and Future Work}
This paper presented a comprehensive framework for manipulator to implement dexterous motion planning task by learning from human demonstration. Through the integration of MediaPipe and depth camera, the framework enables the precise calculation of the 3D coordinates of the human hand, with an error margin of less than $2$ $cm$. Utilizing these coordinates derived from human demonstrations, the framework facilitates the definition and acquisition of position and Euler angles through a modified DMP. This framework not only enhances the robot's capacity to perform various dexterous tasks but also augments its ability to imitate human motion, thereby more flexible and collaborative.

\bibliographystyle{IEEEtran}
\bibliography{IEEEabrv,paper}
\end{document}